\documentclass[journal,comsoc]{IEEEtran}
\usepackage[T1]{fontenc}% optional T1 font encoding
\usepackage{cite}

\ifCLASSINFOpdf
   \usepackage[pdftex]{graphicx}
\else
   \usepackage[dvips]{graphicx}
\fi

\usepackage{amsmath}
\usepackage[cmintegrals]{newtxmath}

\usepackage[numbers]{natbib}
\usepackage{booktabs}
\usepackage{multirow}
\usepackage{multicol}
\usepackage{amsmath}
\usepackage{algpseudocode}
\usepackage{mathtools}
\usepackage[linesnumbered,ruled,vlined]{algorithm2e}
\usepackage{url}

\usepackage{tikz}
\usetikzlibrary{positioning, shapes, arrows.meta, calc}

\mathchardef\mhyphen="2D

\begin{document}
%
% paper title
% Titles are generally capitalized except for words such as a, an, and, as,
% at, but, by, for, in, nor, of, on, or, the, to and up, which are usually
% not capitalized unless they are the first or last word of the title.
% Linebreaks \\ can be used within to get better formatting as desired.
% Do not put math or special symbols in the title.
% \title{Scope expansion for LLM to increase the probability of the correctness of output}
\title{Method Decoration (DeMe): A Framework for LLM-Driven Adaptive Method Generation in Dynamic IoT Environments}

\author{Hong~Su
% <-this % stops a space
\IEEEcompsocitemizethanks{\IEEEcompsocthanksitem H. Su is with the School of Computer Science, Chengdu University of Information Technology, Chengdu, China.\\
E-mail: suguest@126.com. \\
\protect\\
% note need leading \protect in front of \\ to get a newline within \thanks as
% \\ is fragile and will error, could use \hfil\break instead.
}% <-this % stops an unwanted space
\thanks{}}

% The paper headers
\markboth{Journal of \LaTeX\ Class Files,~Vol.~14, No.~8, August~2015}%
{Shell \MakeLowercase{\textit{et al.}}: Bare Demo of IEEEtran.cls for IEEE Communications Society Journals}
% The only time the second header will appear is for the odd numbered pages
% after the title page when using the twoside option.
%
% * Note that you probably will NOT want to include the author's *
% * name in the headers of peer review papers.                   *
% You can use \IFCLASSOPTIONpeerreview for conditional compilation here if
% you desire.

% make the title area
\maketitle

\begin{abstract}
Intelligent IoT systems increasingly rely on large language models (LLMs) to generate task-execution methods for dynamic environments. However, existing approaches lack the ability to systematically produce new methods when facing previously unseen situations, and they often depend on fixed, device-specific logic that cannot adapt to changing environmental conditions.
In this paper, we propose \emph{Method Decoration} (DeMe), a general framework that modifies the method-generation path of an LLM using explicit decorations derived from hidden goals, accumulated learned methods, and environmental feedback. Unlike traditional rule augmentation, decorations in DeMe are not hardcoded; instead, they are extracted from universal behavioral principles, experience, and observed environmental differences. DeMe enables the agent to reshuffle the structure of its method path—through pre-decoration, post-decoration, intermediate-step modification, and step insertion—thereby producing context-aware, safety-aligned, and environment-adaptive methods.
Experimental results show that method decoration allows IoT devices to derive ore appropriate methods when confronting unknown or faulty operating conditions.
\end{abstract}

% Note that keywords are not normally used for peerreview papers.

\IEEEpeerreviewmaketitle

\section{Introduction}

Large language models (LLMs) have recently emerged as powerful reasoning and decision-making components across a wide range of intelligent systems \cite{huang2025foundation} \cite{gholami2024artificial}. In the context of the Internet of Things (IoT), LLMs are increasingly embedded directly into devices to support autonomous operation, local decision-making, and real-time responses without relying on continuous cloud connectivity.
Once deployed, however, these embedded LLMs are typically fixed and difficult to update due to device constraints, limited network access, or security and certification requirements. As a result, IoT devices equipped with pre-installed LLMs often struggle when facing previously unseen situations, environmental variations, or operational faults that were not covered during their initial training. This limitation leads to incorrect or incomplete reasoning, making it difficult for such devices to generate appropriate methods for new or evolving tasks.

Several recent studies have explored how IoT systems can gradually accumulate new experience \cite{su2025method} and derive improved operational strategies through continuous interaction with the environment. However, these approaches do not provide a systematic mechanism for generating new methods based on the acquired experience. Without the ability to explicitly incorporate learned knowledge into the reasoning chain, embedded LLMs remain constrained by their original inference pathways, limiting their capacity to adapt to novel or complex situations.

To address these challenges, this paper proposes \emph{Method Decoration} (DeMe), a general framework that enables LLM-driven IoT devices to adaptively reshape both their task-execution steps and method-generation processes. Rather than replacing or fine-tuning the embedded LLM, DeMe augments the LLM's method path through \emph{decorations}—structured additions that contribute contextual knowledge, intermediate steps, or learned operational rules. These decorations are derived from experience accumulated during device–environment interaction. By integrating the original input with newly acquired decorations, the system can generate improved methods without modifying the LLM's internal weights.

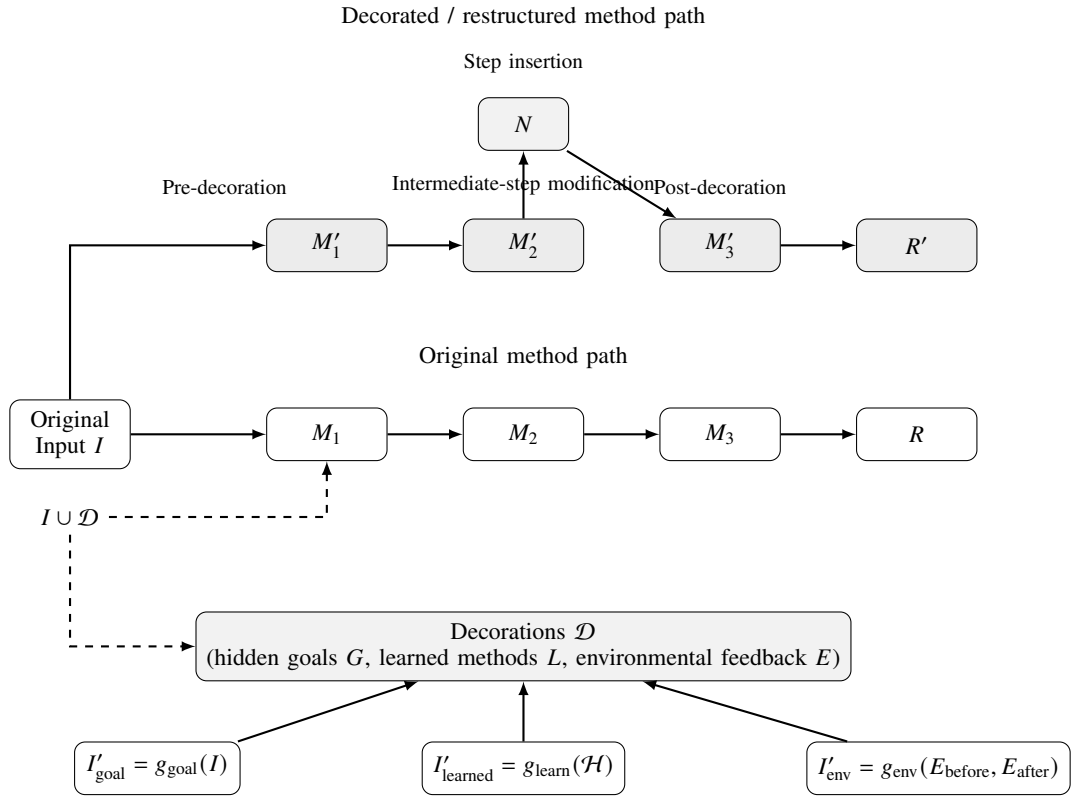
\begin{figure*}[htbp]
    \centering
    \begin{tikzpicture}[
        node distance=10mm and 14mm,
        every node/.style={font=\small},
        box/.style={draw, rounded corners, align=center, inner sep=4pt, minimum width=16mm, minimum height=7mm},
        deco/.style={draw, rounded corners, align=center, inner sep=4pt, minimum width=24mm, minimum height=8mm, fill=gray!10},
        arrow/.style={-latex, thick},
        dashedarrow/.style={-latex, thick, dashed}
    ]

    % Original input
    \node[box] (input) {Original\\Input $I$};

    % Original method path nodes
    \node[box, right=18mm of input] (m1) {$M_1$};
    \node[box, right=10mm of m1] (m2) {$M_2$};
    \node[box, right=10mm of m2] (m3) {$M_3$};
    \node[box, right=10mm of m3] (r)  {$R$};

    % Arrows for original path
    \draw[arrow] (input) -- (m1);
    \draw[arrow] (m1) -- (m2);
    \draw[arrow] (m2) -- (m3);
    \draw[arrow] (m3) -- (r);

    % Label for original path
    \node[align=center, above=4mm of m2] (origlabel) {Original method path};

    % Decoration block (G, L, E)
    \node[deco, below=20mm of m2, minimum width=60mm] (deco) {Decorations $\mathcal{D}$\\
        (hidden goals $G$, learned methods $L$, environmental feedback $E$)};

    % Individual decoration sources
    \node[box, below left=8mm and -6mm of deco, minimum width=22mm] (g) {$I'_{\text{goal}} = g_{\text{goal}}(I)$};
    \node[box, below=8mm of deco, minimum width=22mm] (l) {$I'_{\text{learned}} = g_{\text{learn}}(\mathcal{H})$};
    \node[box, below right=8mm and -6mm of deco, minimum width=32mm] (e) {$I'_{\text{env}} = g_{\text{env}}(E_{\text{before}}, E_{\text{after}})$};

    \draw[arrow] (g) -- (deco);
    \draw[arrow] (l) -- (deco);
    \draw[arrow] (e) -- (deco);

    % Arrow from decorations to decorated path
    \node[below=4mm of input] (idunion) {$I \cup \mathcal{D}$};
    \draw[dashedarrow] (idunion) |- (deco);
    \draw[dashedarrow] (idunion) -| (m1);

    % Decorated path nodes (primed)
    \node[box, above=18mm of m1, fill=gray!15] (m1p) {$M_1'$};
    \node[box, above=18mm of m2, fill=gray!15] (m2p) {$M_2'$};
    \node[box, above=18mm of m3, fill=gray!15] (m3p) {$M_3'$};
    \node[box, above=18mm of r,  fill=gray!15] (rp)  {$R'$};

    % Inserted step N
    \node[box, above=9mm of m2p, fill=gray!10, minimum width=12mm] (nstep) {$N$};

    % Arrows for decorated path
    \draw[arrow] (input.north) |- (m1p);
    \draw[arrow] (m1p) -- (m2p);
    \draw[arrow] (m2p) -- (nstep);
    \draw[arrow] (nstep) -- (m3p);
    \draw[arrow] (m3p) -- (rp);

    % Labels for decoration types
    \node[align=left, above left=2mm and -4mm of m1p] {\footnotesize Pre-decoration};
    \node[align=left, above=2mm of m2p] {\footnotesize Intermediate-step modification};
    \node[align=left, above=2mm of nstep] {\footnotesize Step insertion};
    \node[align=left, above=2mm of m3p] {\footnotesize Post-decoration};

    % Label for decorated path
    \node[align=center, above=8mm of nstep] (declabel) {Decorated / restructured method path};

    \end{tikzpicture}
    \caption{Overview of the proposed Method Decoration (DeMe) framework. Starting from an original
    method path composed of task-execution steps $M_1 \rightarrow M_2 \rightarrow M_3 \rightarrow R$,
    DeMe introduces decorations derived from hidden goals, learned methods, and environmental feedback.
    These decorations enable pre-decoration, intermediate-step modification, step insertion, and
    post-decoration, yielding a restructured path $M_1' \rightarrow M_2' \rightarrow N \rightarrow M_3'
    \rightarrow R'$ that is better aligned with safety requirements and environmental conditions.}
    \label{fig:deme_overview}
\end{figure*}

This paper formalizes two major categories of decorations. \emph{Whole-process decoration} enriches the original input with accumulated contextual knowledge, such as safety-related observations or newly discovered device capabilities. \emph{Step-level decoration}, in contrast, inserts additional intermediate reasoning steps—such as verification or constraint-checking processes—that restructure the reasoning path itself. Together, these mechanisms enable IoT devices to generalize more effectively to unseen conditions and to enhance the robustness of their operational decisions.

The main contributions of this paper are summarized as follows:
\begin{itemize}
    \item We propose DeMe, a new framework that reshapes the method-generation path of embedded LLMs through structured decorations, enabling adaptation without model fine-tuning.
    \item We formalize two major categories of decorations—whole-process decoration and step-level decoration—and illustrate how they modify the reasoning pathway.
    \item We provide a detailed mechanism for learning, recording, and reusing experience-derived decorations in independent IoT systems.
\end{itemize}

The remainder of the paper is organized as follows. Section~\ref{sec:related_work} reviews related research on LLM-driven IoT systems and adaptive reasoning. Section~\ref{sec:model} introduces the proposed DeMe framework. Section~\ref{sec_verification} presents experimental validation across two representative IoT scenarios. Section~\ref{sec:conclusion} concludes the paper and discusses directions for future work.

\section{Related Work} \label{sec:related_work}
Research related to this work spans several areas, including the use of large language models in IoT
and edge environments, adaptive reasoning mechanisms for LLMs, environment-driven learning in IoT
systems, and alternative techniques for modifying model behavior such as fine-tuning, retrieval
augmentation, and prompt engineering. This section reviews the most relevant studies in each of these
domains and highlights their limitations in the context of embedded IoT devices, thereby motivating
the development of the proposed DeMe framework.

\subsection{LLMs in IoT and Edge Intelligence}

Recent advances in large language models (LLMs) have enabled their deployment in increasingly
resource-constrained environments, allowing IoT devices and edge systems to perform autonomous
reasoning and decision-making without continuous cloud connectivity. Several works have explored the
use of compact or distilled LLMs on microcontrollers, edge gateways, and embedded platforms
\cite{qu2025mobile,yao2025efficient}. These studies demonstrate that LLM-enabled
IoT devices can interpret sensor inputs, execute multi-step tasks, and coordinate distributed
components in real time.

However, embedding an LLM within an IoT device introduces strict limitations. Once deployed, the
model is often difficult or impossible to update due to restricted compute resources, long-term
certification requirements, or the absence of reliable connectivity \cite{zong2025integrating}. As a
result, the embedded LLM must operate for long periods in environments that evolve beyond the
conditions represented in its training data. Prior works highlight that such static LLMs tend to
generate incorrect or incomplete responses when facing previously unseen operational states,
environment faults, or unusual contextual variations \cite{sarhaddi2025llms}. These
limitations make traditional fine-tuning or cloud-assisted adaptation impractical for independent
IoT devices.

While the above frameworks illustrate how LLMs can be optimized for on-device inference, they focus primarily on model compression and do not address the fundamental challenge of improving reasoning quality after deployment. To date, no existing work provides a mechanism for modifying or enriching the LLM's method-generation process without retraining the underlying model. This gap motivates the development of the Method Decoration (DeMe), which aims to enhance adaptability in embedded IoT LLMs without modifying their internal parameters.

\subsection{Adaptive Reasoning and Method Generation in LLMs}

LLMs have shown strong capabilities in structured reasoning, step-by-step planning, and generating
multi-stage solution methods. Techniques such as chain-of-thought prompting \cite{wei2022chain},
self-consistency decoding \cite{wang2022self}, and deliberative reasoning
\cite{zhou2022least} enable models to generate intermediate steps that improve the reliability of
final answers. More advanced forms of method generation include self-refinement
\cite{madaan2023self}, ReAct-style reasoning-and-action loops \cite{yao2022react}, tool-augmented
reasoning \cite{schick2023toolformer}, and multi-agent self-collaboration \cite{dong2024self}.

However, these techniques primarily modify the \emph{prompting strategy} rather than the underlying
method-generation process. The LLM follows the same inference pathway, even when the reasoning
approach is altered through instruction patterns. Moreover, most of these methods rely on iterative
feedback loops or external tool APIs, which are impractical for embedded IoT devices that lack full
connectivity or computational resources.

A related line of work investigates how LLMs can revise their reasoning processes by generating
alternate solutions or performing internal verification \cite{kamoi2024can}. While these works
show the potential of LLMs to critique or correct their own output, they do not introduce a mechanism
for systematically \emph{decorating} the reasoning path with new, experience-derived information.
Their refinement processes remain ephemeral and do not accumulate across tasks or over time.

In contrast, the proposed Method Decoration(DeMe) allows a device to modify its
method-generation pathway by explicitly incorporating newly learned methods, contextual knowledge,
and intermediate reasoning steps. Rather than adjusting prompts heuristically, DeMe provides a
structured framework for persistent, reusable, and interpretable modifications to the reasoning
process—an ability absent from existing adaptive reasoning techniques.

\subsection{Learning from Environment Interaction in IoT Systems}

IoT devices frequently operate in dynamic and unpredictable environments, making it necessary for them to adapt their behavior as conditions evolve. A substantial body of work explores how IoT systems learn from environmental feedback. Reinforcement learning (RL) and online policy adaptation are widely used to refine control strategies for energy management, autonomous navigation, and resource allocation \cite{mnih2015human, lillicrap2015continuous, chen2021iotrl}. These approaches rely on trial-and-error interactions, allowing the system to gradually improve its decision-making over time.

Beyond RL, many IoT applications employ rule adaptation or experience-based knowledge refinement \cite{li2020iotadapt, sun2019edgelearning}. Such systems incrementally update operational rules or heuristics when new patterns or environmental states are observed. Recent studies also explore continual learning frameworks for edge devices, which incrementally integrate new information without catastrophic forgetting \cite{parisi2019continual}. These techniques enable long-term adaptation but require model updates, additional memory buffers, or replay strategies that may be infeasible for embedded IoT hardware.

A key limitation of environment-driven learning approaches is that they typically operate \emph{outside} the LLM. They may adjust control parameters, update external rules, or train new lightweight models, but they do not modify the LLM’s internal reasoning strategy or method-generation pathway. As a result, the embedded LLM may continue producing inappropriate or outdated methods even after the device acquires new operational knowledge.

The Method Decoration(DeMe) builds on the idea of accumulating experience from the environment but differs fundamentally in its integration mechanism. Instead of updating the model or replacing rules, DeMe records learned methods explicitly and reinjects them as structured decorations that reshape the LLM’s reasoning process. This allows embedded IoT systems to benefit from adaptive experience-based improvements while avoiding the computational and security challenges associated with continual model updates.

\subsection{Model Modification Alternatives: Fine-Tuning, RAG, and Prompt Engineering}

Several approaches have been proposed to adapt LLM behavior without retraining from scratch. One
widely used strategy is parameter-based fine-tuning, including full-model fine-tuning
\cite{ouyang2022instructgpt}, Low-Rank Adaptation (LoRA) \cite{hu2022lora}, and other lightweight
parameter-efficient techniques \cite{lester2021prompttuning,zaken2022bitfit}. While effective in many
cloud or server-side applications, these methods require additional training cycles, gradient
updates, and storage of new parameter sets—making them unsuitable for embedded IoT devices, where
models are often non-upgradable due to hardware constraints or certification requirements.

Retrieval-Augmented Generation (RAG) offers a non-parametric alternative by incorporating external
knowledge retrieved from local or remote databases \cite{lewis2020rag, borgeaud2022retro}. RAG
systems can dynamically adapt to new information without changing model parameters. However, RAG
assumes access to a searchable knowledge base and introduces memory, storage, and retrieval
latency challenges that are not acceptable for low-power or intermittently connected IoT devices.
Furthermore, RAG influences the model at the content level but does not alter the method-generation
path or reasoning structure.

Prompt engineering techniques, such as instruction tuning \cite{wei2021finetuned},
task-specific prompt templates \cite{kojima2022zero_shot}, or structured few-shot examples
\cite{brown2020gpt3}, offer another approach for modifying model behavior. While these techniques
are simple to apply, they focus primarily on manipulating textual instructions rather than providing
a systematic mechanism for integrating newly learned methods. Their effects are often unstable,
context-dependent, and difficult to record or reuse consistently across diverse IoT tasks.

In contrast to these alternative adaptation methods, the Method Decoration(DeMe) enables
embedded IoT devices to modify the method-generation process without changing model parameters,
requiring external databases, or depending on carefully engineered prompt patterns. By explicitly
recording and injecting learned decorations into the reasoning chain, DeMe provides a persistent,
interpretable, and computationally lightweight means of enhancing LLM adaptability in resource-
constrained environments.

\section{The Method Decoration (DeMe) Framework}
\label{sec:model}

Existing systems lack a systematic mechanism for generating new methods when previously unencountered situations arise or when improved strategies are required for specific environments. At the same time, current IoT devices typically operate using fixed, device-specific logic rather than a generalizable reasoning process. This limits their adaptability and prevents them from exploiting broader contextual factors such as environmental dynamics, task-related requirements, or implicit safety constraints. In contrast, DeMe leverages these broader considerations as \emph{decorations}, which can be incorporated into the method-generation process—often through a pre-installed LLM—to produce new or improved methods.

To introduce method decoration, we first examine how methods are generated by an LLM under the standard setting. Large language models (LLMs) typically produce a method by mapping an input description $I$ to a sequence of internal reasoning steps that ultimately yield an actionable output~$R$. This process can be abstracted as a direct reasoning path:
\begin{equation}
    I \rightarrow R,
\end{equation}
where $I$ encodes the problem statement and contextual information, and $R$ represents the method generated by the LLM to solve the task.

By incorporating learned experience and environmental feedback, the proposed Method Decoration (DeMe) framework transforms this original single-path formulation into a richer and more expressive process:
\begin{equation}
    (I, I', \mathcal{D}) \rightarrow R',
\end{equation}
where $I'$ represents accumulated experience or previously learned methods, and $\mathcal{D}$ denotes optional decorations such as hidden goals, risk indicators, intermediate-step refinements, or environmental feedback. By explicitly injecting these elements into the reasoning pipeline, DeMe provides a principled mechanism for generating new, context-aware, and environment-aligned methods.

Unless otherwise ambiguous, we do not explicitly distinguish between $I'$ and $\mathcal{D}$. For simplicity of notation, we use $I'$ to represent both.

Method decoration enables us to enhance existing methods so they can meet new or more complex demands as our knowledge grows or the situation changes. When we acquire new information or face unfamiliar challenges without direct guidance, method decoration allows us to adapt and refine our approaches accordingly.

\subsection{Method Path Representation}
\label{subsec:model_path}

To describe how decorations operate within DeMe, we first formalize the notion of a \emph{method path}. A method path represents the sequence of \emph{task-execution steps} that an IoT agent performs to complete an assigned task. A generated method typically consists of multiple actionable steps, and these steps form an ordered path that governs how the task is carried out.

Formally, a method path can be expressed as
\begin{equation}
    \textit{Method}: \quad M_1 \rightarrow M_2 \rightarrow \cdots \rightarrow M_k,
\end{equation}
where $M_i$ denotes the $i$-th reasoning step in the method produced by the LLM.

There may be different ways to obtain such a method path. For example, the two-step method in
\eqref{eq_two_steps} can be generated in a single LLM invocation, as in \eqref{eq_one_step_gen}, or in
multiple invocations, as in \eqref{eq_two_step_gen}. For simplicity, we do not distinguish these
generation modes in this paper.
\begin{equation} \label{eq_two_steps}
    \textit{Method}: \quad M_1 \rightarrow M_2,
\end{equation}
\begin{equation} \label{eq_one_step_gen}
    I' \xrightarrow{\text{LLM}} (M_1 \rightarrow M_2),
\end{equation}
\begin{equation} \label{eq_two_step_gen}
    I' \xrightarrow{\text{LLM}} M_1 \xrightarrow{\text{LLM}} M_2.
\end{equation}

With the method-path perspective, decorations can be applied to individual reasoning stages rather
than only to the entire method. Examples include:
\begin{itemize}
    \item Adding constraints or contextual knowledge at the \textbf{beginning} to modify the entry point of the path.
    \item Refining or adjusting internal reasoning steps to modify the \textbf{middle portion} of the path.
    \item Appending corrective logic after the original output to modify the \textbf{end} of the path.
\end{itemize}

Thus, a method can be improved not only by modifying its original input or final output as a whole, but also by restructuring its intermediate task-execution steps. This leads to the decorated-path representation:

\begin{equation}
    (I, I') \rightarrow M_1' \rightarrow M_2' \rightarrow \cdots \rightarrow M_k',
\end{equation}
or, when decoration is applied to a specific internal step,
\begin{equation}
    I \rightarrow \cdots \rightarrow (M_\ell, I') \rightarrow M_{\ell+1}' \rightarrow \cdots \rightarrow M_k',
\end{equation}
where $I'$ denotes accumulated experience or additional knowledge, and $M_i'$ represents the modified
reasoning steps after decoration.

% This method-path viewpoint provides the foundation for the decoration categories introduced in the following subsection and explains how DeMe systematically reshapes reasoning behavior by operating on specific regions of the method path.

\subsection{Types of Method Decoration}
\label{subsec:model_types}

The Method Decoration (DeMe) framework modifies the task-execution path by introducing additional
information, refinements, or structural adjustments. These modifications, referred to as
\emph{decorations}, systematically reshape how an LLM constructs or updates the operational steps of a
method. Based on how the task-execution path is altered, we categorize decorations into four major
types: pre-decoration, post-decoration, intermediate-step modification, and step insertion.

\subsubsection{Pre-decoration}

Pre-decoration adds supplementary information directly to the method-generation input, transforming
the basic mapping $I \rightarrow M$ into a decorated mapping $(I, I') \rightarrow M'$, where $I'$
encodes additional knowledge, constraints, or safety rules. Because the LLM incorporates $I'$ before
producing the first task-execution step, the entire method path is influenced.

Formally,
\begin{equation}
    \{I, I'\} \xrightarrow{\text{LLM}} M'.
\end{equation}
Pre-decoration is effective when the method must consistently account for additional context, such as
hidden goals, device limitations, or safety protocols.

\subsubsection{Post-decoration}

Post-decoration modifies the method after the original procedure has been generated. Rather than
altering the task-execution path during generation, this approach refines or augments the final
output:
\begin{equation}
    I \xrightarrow{\text{LLM}} M, \qquad (M, I') \rightarrow \hat{M}.
\end{equation}
Post-decoration is suitable when the original method remains largely correct but requires additional
checks, thresholds, or corrective rules—for example, adding a review step to validate the feasibility
of the final action.

\subsubsection{Intermediate-step Modification}

Intermediate-step modification targets specific operational steps within the method path. Given an
original process
\begin{equation}
    I \rightarrow M_{a} \rightarrow M_{b} \rightarrow M_c \rightarrow M_d,
\end{equation}
DeMe selectively replaces one or more intermediate steps with refined versions:
\begin{equation}
    I \rightarrow (M_{a}, I') \rightarrow M_{b}' \rightarrow M_{c}' \rightarrow M_{d}'.
\end{equation}
This decoration is useful when most task steps are correct but certain sub-decisions require
adjustment—such as correcting an unsafe actuation step while keeping the sensing and validation
steps unchanged.

\subsubsection{Step Insertion}

Step insertion extends the method path by adding one or more new task-execution steps. The original
path is
\begin{equation} \label{eq_ste_ins_ori}
    I \rightarrow M_a \rightarrow M_b,
\end{equation}
and the decorated version becomes
\begin{equation} \label{eq_ste_ins_aft}
    I \rightarrow M_n \rightarrow M_a' \rightarrow M_b',
\end{equation}
where $M_n$ performs additional operations such as prediction, constraint checking, or verification
before the main method step is executed. Step insertion is particularly effective in safety-critical
or dynamically changing environments because it enables early detection of potential constraint
violations or anomalies.

\medskip

Collectively, these four forms of decorations offer a flexible and systematic way to adjust and
enhance the operational structure of a method. They enable fine-grained control over different
segments of the task-execution path, while retaining interpretability and allowing explicit
integration of learned experience and environmental feedback.

\subsection{Decoration Generation Mechanism}
\label{subsec:model_decoration_generation}

Decorations play a central role in DeMe by providing the additional information or structural
guidance required to reshape the task-execution path. In this subsection, we formalize how
decorations are generated, recorded, and incorporated into the method-generation process. The
proposed mechanism draws from three major sources: hidden goals, accumulated learned methods, and
environmental feedback.

\subsubsection{Hidden Goals} \label{sec_hid_goa}

Hidden goals represent implicit expectations or universal principles that an IoT agent should follow,
even when they are not explicitly included in the input. Examples include ensuring safety, avoiding
unnecessary risk, conserving resources, or completing tasks thoroughly. Such goals frequently arise
in IoT applications but may be omitted from the initial prompt provided to the LLM.

To make these goals available during method generation, DeMe records them explicitly as decorations:
\begin{equation}
    I'_{\text{goal}} = \text{HiddenIntent}(I),
\end{equation}
where $I'_{\text{goal}}$ is merged with the agent's input during method generation. This ensures that
the resulting method path reflects these overarching principles consistently.

\subsubsection{Accumulated Learned Methods} \label{sec_acc_lea_}

IoT agents often operate in dynamic and evolving environments. During interaction, the agent may
encounter scenarios that require alternative strategies or refined task-execution steps. When an
improved method is discovered—because it performs better or successfully handles a previously
unsolved case—it is recorded as experience:
\begin{equation}
    I'_{\text{learned}} = \text{RecordBetterMethod}(M).
\end{equation}

This accumulated experience becomes part of the decorated input used in subsequent method generation:
\begin{equation}
    (I, I'_{\text{learned}}) \rightarrow M'.
\end{equation}
Such incremental accumulation is consistent with the progressive method-learning mechanism described
in \cite{su2025method}, enabling the agent to evolve its method library over time.

\subsubsection{Environmental Feedback} \label{sec_env_fed}

Environmental feedback provides direct evidence of how a generated method influences the surrounding
system. Unlike fixed measurements or predefined metrics, this feedback captures general environmental
states before, during, and after the method is executed. Let $E_{\text{before}}$ and $E_{\text{after}}$
be the environmental states immediately before and after executing the method. The difference
\begin{equation} \label{eq_env_diff}
    \Delta E = E_{\text{after}} - E_{\text{before}}
\end{equation}
reveals the effect of the method. DeMe extracts this information to form a decoration:
\begin{equation}
    I'_{\text{env}} = \text{ExtractImpact}(\Delta E).
\end{equation}

This idea aligns with the difference-based reasoning mechanism introduced in
\cite{su2025difference}, where changes across temporal states help identify meaningful impacts and
enable the agent to refine future task-execution steps.

\subsubsection{Combined Decoration Input}

All decoration components are explicitly recorded, rather than stored implicitly within the LLM’s
internal parameters. The combined decoration set is defined as
\begin{equation}
    \mathcal{D} = \{ I'_{\text{goal}}, I'_{\text{learned}}, I'_{\text{env}} \}.
\end{equation}

The decorated method-generation process is then expressed as
\begin{equation}
    (I, \mathcal{D}) \rightarrow M',
\end{equation}
where $M'$ denotes the refined method generated under the influence of all decorations.

This explicit and structured mechanism allows DeMe to incorporate safety requirements, accumulated
operational experience, and environmental impact information into the task-generation procedure,
enabling the production of adaptive, robust, and context-aware methods across diverse IoT scenarios.

\subsection{Why Decorations Enable Improved Methods: A Brief Proof Sketch}
\label{subsec:proof_decorations}

The DeMe framework assumes that improved methods arise from three key sources of additional
information: hidden goals, accumulated learned methods, and environmental feedback. In this
subsection, we provide a brief proof sketch showing that these decorations are \emph{necessary} for
producing improved methods, and that without them, an LLM relying solely on the original input $I$
cannot guarantee better solutions.

\subsubsection{Incomplete Information in the Original Input}

Let $M = f_{\text{LLM}}(I)$ denote the method generated by an LLM from the input~$I$.
The input $I$ typically describes only the immediate task requirements and lacks the following
essential components:

\begin{itemize}
    \item \textbf{Implicit behavioral principles (hidden goals).}
    These represent normative constraints that must hold for any acceptable method, such as safety,
    completeness, or rule compliance. Hidden goals are not encoded in $I$, but instead correspond to
    additional constraints of the form
    \begin{equation}
        G = g_{\text{goal}}(I),
    \end{equation}
    where $G$ captures expectations that extend beyond the explicit task description.

    \item \textbf{Knowledge of previously successful methods (learned methods).}
    Improvement through experience, as described in~\cite{su2025method}, requires retaining and
    reusing past high-quality methods. Let $\mathcal{H}$ denote the history of previously executed
    methods and outcomes. The corresponding learned knowledge is
    \begin{equation}
        L = g_{\text{learn}}(\mathcal{H}),
    \end{equation}
    which is not contained in the original input~$I$.

    \item \textbf{Observed effects of past executions (environmental feedback).}
    The environment may change dynamically and is not represented by fixed measurements. Instead,
    meaningful feedback is extracted from the difference between environmental states before and
    after executing a method (see Section~\ref{sec_env_fed}):
    \begin{equation}
        E = g_{\text{env}}\!\left(E_{\text{after}} - E_{\text{before}}\right).
    \end{equation}
    Such feedback identifies impacted factors and indicates whether refinement is required.
\end{itemize}

Thus, the LLM’s mapping $I \rightarrow M$ is based on a limited information set.
Formally, let $\mathcal{I}_0$ denote the information contained in $I$.
Any improved method $M^\star$ that depends on information outside $\mathcal{I}_0$ cannot be generated
unless that additional information is explicitly supplied. Hence, if $M^\star$ requires information
not encoded in $\mathcal{I}_0$, we have
\begin{equation} \label{eq_mstar_not_m}
    M^\star \notin f_{\text{LLM}}(\mathcal{I}_0).
\end{equation}

\subsubsection{Decorations Expand the Effective Information Set}

Decorations introduce additional information
\begin{equation}
    \mathcal{D} = \{ I'_{\text{goal}},\, I'_{\text{learned}},\, I'_{\text{env}} \},
\end{equation}
resulting in an expanded input set
\begin{equation}
    \mathcal{I}_1 = \mathcal{I}_0 \cup \mathcal{D}.
\end{equation}

Because $\mathcal{D}$ contains information unavailable in $I$, we have
\begin{equation}
    \mathcal{I}_1 \supset \mathcal{I}_0.
\end{equation}

These decorations are not hardcoded. Each component is generated by a general
function rather than a device-specific constant:
\begin{equation}
\begin{aligned}
    I'_{\text{goal}}    &= g_{\text{goal}}(I), \\
    I'_{\text{learned}} &= g_{\text{learn}}(\mathcal{H}), \\
    I'_{\text{env}}     &= g_{\text{env}}(E_{\text{before}},\, E_{\text{after}}).
\end{aligned}
\end{equation}

where $g_{\text{goal}}$, $g_{\text{learn}}$, and $g_{\text{env}}$ are task-independent mappings, and
$\mathcal{H}$ denotes the history of past methods. Since these decorations are not fixed constants,
\begin{equation}
    I'_{\text{goal}},\, I'_{\text{learned}},\, I'_{\text{env}} \notin \mathrm{Const},
\end{equation}
they automatically adapt across tasks, environments, and devices.

Therefore, an improved method $M'$ becomes reachable as long as the necessary information lies within
$\mathcal{D}$:
\begin{equation}
    M' \in f_{\text{LLM}}(\mathcal{I}_1)
    \qquad \text{while} \qquad
    M' \notin f_{\text{LLM}}(\mathcal{I}_0).
\end{equation}

\subsubsection{Decorations Provide Corrective Direction for Improvement}

Let $G$ denote the hidden goals,
$L$ the accumulated learned methods,
and $E$ the environmental feedback signals.
Let improvement be measured by a performance metric $J(\cdot)$ such that
\begin{equation}
    J(M') > J(M).
\end{equation}

Decorations improve methods through the following mechanisms:

\begin{enumerate}
    \item \textbf{Hidden goals provide normative constraints.}
    If $M$ violates $G$, incorporating $I'_{\text{goal}}$ restricts the search to feasible or safe
    methods:
    \begin{equation}
        M' \models G \quad \Rightarrow \quad J(M') > J(M).
    \end{equation}

    \item \textbf{Learned methods provide constructive corrections.}
    If a previously successful method $R_{\text{best}}$ exists,
    \begin{equation}
        J(R_{\text{best}}) > J(M),
    \end{equation}
    including $I'_{\text{learned}}$ biases the LLM toward refinements that inherit its effective
    components.

    \item \textbf{Environmental feedback provides empirical adjustments.}
    If execution of $M$ yields environmental change $\Delta E$, then
    \begin{equation}
        I'_{\text{env}} = \text{ExtractImpact}(\Delta E)
    \end{equation}
    provides a corrective signal indicating which task-execution steps require modification.
\end{enumerate}

Collectively,
\begin{equation} \label{eq_all_fac}
    \mathcal{D} = \{G, L, E\}
\end{equation}
serves as a sufficient condition for improvement.
\\
\par

Decorations expand the LLM’s effective input space and supply the normative, experiential, and
environmental structure required to improve task-execution methods. Because they are generated
automatically from universal principles and observations rather than hardcoded rules, decorations are
general, adaptive, and transferable across IoT tasks. Without them, improved methods may be
inaccessible or indistinguishable from suboptimal ones, making decorations a necessary and enabling
component of the DeMe framework.

\section{Verification}
\label{sec_verification}

In this section, we evaluate the effectiveness of the proposed Method Decoration(DeMe) across two IoT-oriented scenarios. The first scenario verifies whole-process decoration, where the entire reasoning path is modified using additional contextual knowledge. The second scenario examines step-insertion decoration within a smart-building HVAC control system.

\subsection{Decoration for the Whole Process}
\label{subsec:whole_process}

This experiment evaluates whether DeMe can modify the \emph{entire} reasoning process by extending the original path $I \rightarrow M$ into a decorated path $(I, I') \rightarrow M'$. The objective is to test whether adding contextual knowledge as decoration enables an LLM to generate methods that better align with an implicit safety goal.

We consider an IoT device analogous to a small autonomous vehicle equipped with a throttle and a brake. The operational description below is provided to the LLM to establish a consistent baseline understanding of the device’s capabilities:

\begin{quote}
``This IoT device has a throttle and a brake. To accelerate, adjust the throttle button. The throttle has four different levels, corresponding to different acceleration forces. Level~1 provides the smallest acceleration, while Level~4 provides the strongest acceleration. The brake has four different levels, corresponding to different deceleration forces. When the speed reaches the preset value, maintain the current operation. If neither acceleration nor deceleration is needed, engage the brake to prepare for unexpected situations.''
\end{quote}

The test scenario asks the LLLM to describe how to operate the device on rainy or snowy days when the
\emph{primary brake is broken}. Two prompting strategies are evaluated:

\begin{itemize}
    \item \textbf{M\textsubscript{direct}:}
    The LLM receives only the original prompt $I$:
    \begin{quote}
    ``How to operate this IoT device on rainy or snowy days? The brake is broken. Please describe in one paragraph.''
    \end{quote}

    \item \textbf{M\textsubscript{ours}:}
    The LLM receives the decorated prompt $(I, I')$, where $I'$ contains a safety-related fact learned
    from prior interactions—that the device includes a backup brake system:
    \begin{quote}
    ``How to operate this IoT device on rainy or snowy days? The brake is broken. \emph{Notice this IoT
    device has a backup brake system, and you can use the backup brake system.} Please describe in one
    paragraph.''
    \end{quote}
\end{itemize}

This design simulates a realistic IoT scenario in which certain capabilities (e.g., a backup brake) may initially be unknown to the system but become available through accumulated interaction experience, as described in section \ref{sec_acc_lea_}. Following the principles of~\cite{su2025method}, the learned knowledge is explicitly recorded as $I'$ and reinjected into the reasoning path to form the decorated mapping $(I, I')
\rightarrow M'$.

\subsubsection{Experimental Setup}

The experiment proceeds by querying each prompting strategy 20 times using the DeepSeek platform\footnote{\url{https://chat.deepseek.com/}}, accessed on 22 November 2025. Each generated response is evaluated against a manually constructed reference guideline representing a hidden safety
goal.

\paragraph{One Hidden Goal (Safety) as Reference}
To assess whether a generated method reflects appropriate safety behavior, each output is compared to the following reference instruction:

\begin{quote}
``You should use the backup brake or do not operate this device, as having no brake available could
cause an accident.''
\end{quote}

This reflects standard safety principles for vehicle-like systems (corresponding to the hidden requirements discussed in section \ref{sec_hid_goa}) and represents the expected correct behavior when the primary brake is malfunctioning.

\paragraph{Evaluation Procedure}
Each response from M\textsubscript{direct} and M\textsubscript{ours} is embedded into a vector
representation and compared with the reference guideline using semantic similarity scoring.
Higher similarity indicates a closer match to safety-oriented reasoning.

\subsubsection{Results and Discussion}

Table~\ref{tab_decoration_similarity} presents the similarity scores for the 20 responses generated
under each prompting condition. The direct method (M\textsubscript{direct}) achieves an average
similarity score of 0.5539, whereas the decorated method (M\textsubscript{ours}) achieves a higher
average of 0.6554. This represents an absolute improvement of 0.1015 (approximately 18.3\%
relative gain), indicating that whole-process decoration substantially enhances the safety alignment
of LLM-generated instructions.

\begin{table}[htbp]
  \centering
  \caption{Similarity results for the direct and decorated methods over 20 test rounds.}
  \label{tab_decoration_similarity}
  \begin{tabular}{ccc}
    \toprule
    Episode & Direct (M\textsubscript{direct}) & Ours (M\textsubscript{ours}) \\
    \midrule
    1 & 0.5420 & 0.6399 \\
    2 & 0.5016 & 0.6682 \\
    3 & 0.5016 & 0.6399 \\
    4 & 0.5015 & 0.6399 \\
    5 & 0.5614 & 0.6590 \\
    6 & 0.5755 & 0.6399 \\
    7 & 0.5205 & 0.6399 \\
    8 & 0.5766 & 0.6430 \\
    9 & 0.5542 & 0.6620 \\
    10 & 0.5666 & 0.6711 \\
    11 & 0.5755 & 0.6710 \\
    12 & 0.5631 & 0.6631 \\
    13 & 0.5442 & 0.6430 \\
    14 & 0.5722 & 0.6710 \\
    15 & 0.5547 & 0.6710 \\
    16 & 0.5875 & 0.6430 \\
    17 & 0.5862 & 0.6710 \\
    18 & 0.5603 & 0.6589 \\
    19 & 0.5666 & 0.6413 \\
    20 & 0.5666 & 0.6710 \\
    \midrule
    Mean & \textbf{0.5539} & \textbf{0.6554} \\
    \bottomrule
  \end{tabular}
\end{table}

\begin{figure*}[htbp]
    \centering
    \includegraphics[width=5.5in]{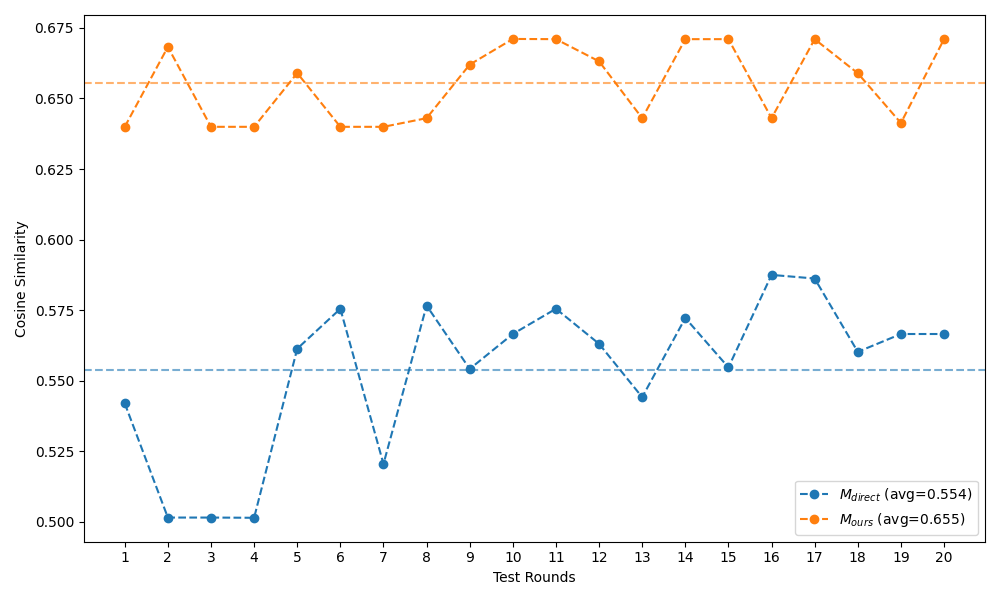}
    \caption{Similarity comparison between the direct prompts (M\textsubscript{direct})
    and the decorated prompts with added safety knowledge (M\textsubscript{ours}).}
    \label{fig_decoration}
\end{figure*}

Figure~\ref{fig_decoration} visualizes these results. The responses generated by
M\textsubscript{direct} exhibit noticeable variability and frequently omit essential safety
considerations, such as relying on the backup brake or recommending halting device operation in
dangerous conditions. In contrast, the decorated method (M\textsubscript{ours}) consistently
produces outputs that align closely with the reference guideline.

Two important observations emerge from the data:

\begin{itemize}
    \item \textbf{Reduced variance:}
    The direct method exhibits a wide range of similarity values (0.5015--0.5875), indicating inconsistent decision-making with respect to the default safety goal. In contrast, the decorated method yields similarity values that are not only higher but also more tightly clustered (0.6399--0.6711), demonstrating far more stable and reliable reasoning.

    \item \textbf{Consistent safety improvement:}
    For all 20 episodes, the decorated prompt yields higher similarity scores than the direct prompt,
    with per-episode gains ranging from 0.07 to 0.17. This confirms that the inclusion of additional
    contextual knowledge ($I'$) systematically enhances the LLM’s ability to identify and prioritize
    safe behavior.
\end{itemize}

These results validate the central principle of DeMe:
\emph{decorating the entire input with accumulated or previously learned safety-related information
reshapes the LLM’s internal reasoning in a predictable and beneficial manner}.
Unlike simple post-processing or rule filtering, DeMe modifies the reasoning path itself, leading to
more robust and interpretable safety behavior in IoT control scenarios.

\subsection{Introducing Additional Steps in DeMe}
\label{subsec:step_insertion}

While the previous experiment evaluates whole-process decoration by modifying the input to the reasoning path, this case study investigates a different form of decoration: the insertion of additional reasoning steps. In DeMe, step insertion allows the agent to introduce new intermediate operations—such as verification, prediction, or constraint evaluation—before generating the final method.

\textbf{Simulation configuration.}
To evaluate this mechanism, we employ a lightweight HVAC (heating, ventilation, and air conditioning) simulation environment that captures indoor thermodynamic evolution, outdoor temperature cycles, and low-cardinality occupancy. Occupancy is observed and interpreted through the environmental-difference comparison defined in~\eqref{eq_env_diff} of Section~\ref{sec_env_fed}.
The key simulation parameters are summarized as follows:

\begin{itemize}
    \item \textbf{Episode length}: 96 steps per day (15 minutes per control step).
    \item \textbf{Indoor temperature dynamics}:
        indoor temperature evolves according to outdoor temperature difference, occupant heat gain (0--3 persons), HVAC cooling effect, and small stochastic noise.
    \item \textbf{Outdoor temperature}:
        sinusoidal daily cycle with mean $30^\circ$C and amplitude $6^\circ$C.
    \item \textbf{Occupancy}:
        randomly sampled each step with distribution $\{0:40\%, 1:30\%, 2:20\%, 3:10\%\}$.
    \item \textbf{HVAC power range}:
        normalized action $a \in [-1,1]$ mapped to actual cooling power in $[0,5]$ kW.
    \item \textbf{Energy consumption}:
        per-step usage computed as $E_t = \mathrm{Power}_t \times 0.25$~kWh (15-minute interval).
    \item \textbf{Anomaly threshold}:
        ``no-person but high-power'' defined when occupancy $=0$ and power $>0.8$~kW.
\end{itemize}

The baseline control logic follows the original reasoning path
\begin{equation}
    I \rightarrow B \rightarrow R,
\end{equation}
where the controller $B$ generates HVAC actions directly from the building’s environmental state $I$ (including indoor temperature and time), without any explicit occupancy-aware verification. This structure reflects conventional rule-based or model-based control without internal safety checks.

In contrast, the decorated controller augments the reasoning path by inserting a verification step $N$:
\begin{equation}
    \{I, I'\} \rightarrow N \rightarrow B' \rightarrow R'.
\end{equation}
Here, $I'$ encodes occupancy and energy-related constraints, and $N$ is responsible for detecting anomalous configurations such as “no person but high HVAC power.’’ When such an anomaly is predicted, $N$ triggers the generation of a refined strategy $B'$ that reduces HVAC power instead of directly applying the baseline controller $B$. This modification alters the method-generation path itself, aligning with the DeMe principle of restructuring reasoning rather than post-correcting results.

Both the baseline and decorated controllers are evaluated over 20 independent episodes, each representing one full day of operation (96 control steps). The next subsection describes the evaluation metrics used in this study, followed by detailed experimental results and analysis.

\subsubsection{Evaluation Metrics}

To compare the baseline controller and the decorated controller with the inserted verification step $N$, we evaluate their performance using four metrics: temperature tracking error (during occupied periods), total energy consumption, wasted energy in unoccupied states, and anomaly counts. These metrics capture both control performance and anomaly-awareness, providing a comprehensive assessment of the effect of step-insertion decoration.

\paragraph{Average Temperature Error in Occupied Periods}

The temperature error measures how closely the indoor temperature $T_t$ tracks the desired setpoint
$T^{*}$ during time steps when the space is occupied. Let $\mathcal{T}_{\text{occ}}$ denote the set of time indices with nonzero occupancy. The average error is defined as
\begin{equation}
    \text{ErrOcc} =
    \frac{1}{|\mathcal{T}_{\text{occ}}|}
    \sum_{t \in \mathcal{T}_{\text{occ}}} \left| T_t - T^{*} \right|.
\end{equation}
A lower ErrOcc indicates tighter temperature control when occupants are present.

\paragraph{Total Energy Consumption}

Total energy consumption reflects the overall electrical energy required to operate the HVAC system during an episode:
\begin{equation}
    E_{\text{total}} = \sum_{t=1}^{T} E_t,
\end{equation}
where $E_t$ denotes the instantaneous energy usage at time $t$ (converted to kWh per simulation step). This metric is essential for evaluating the operational efficiency of the controller.

\paragraph{Wasted Energy in Unoccupied States}

To explicitly capture energy waste, we measure the energy used when the room is unoccupied. Let $\mathrm{Occ}_t$ denote the occupancy at time $t$ and $\mathbb{1}(\cdot)$ the indicator function. The wasted energy is defined as
\begin{equation}
    E_{\text{waste}} =
    \sum_{t=1}^{T}
    \mathbb{1}\!\left( \mathrm{Occ}_t = 0 \right) \, E_t.
\end{equation}
This metric sums the HVAC energy usage during all time steps with zero occupancy.
A lower $E_{\text{waste}}$ reflects more occupancy-aware and energy-efficient control behavior.

\paragraph{Anomaly Counts}

Finally, we quantify anomalous behaviors of the controller. We define an \emph{environment anomaly} (EnvAnom) at time $t$ when the room is unoccupied yet the HVAC power exceeds a given threshold, i.e.,
\emph{``no person but high power''}. For the decorated controller, we also record the number of anomalies intercepted by the inserted step $N$ (N\_Anom), i.e., cases where $N$ detects such a configuration and modifies the control action accordingly. These counts reflect the safety-awareness and corrective capability enabled by step insertion.

\medskip

Together, these metrics provide insight into the trade-offs between anomaly-aware control introduced by step insertion and traditional goals of tracking accuracy and energy efficiency. The next subsection presents the experimental results and discusses the impact of the inserted reasoning step on controller behavior.

\subsubsection{Experimental Results}

Table~\ref{tab:dmom_stepN_episode_compare} summarizes the per-episode performance of the baseline
controller (denoted ``B'') and the decorated controller with step~$N$ (denoted ``D''). Each row reports the occupied-period temperature error (ErrOcc), total energy consumption ($E_{\text{total}}$), wasted energy in unoccupied states ($E_{\text{waste}}$), and anomaly counts for one episode.

Across the 20 episodes, the decorated controller maintains temperature-tracking performance comparable to the baseline. The energy-related metrics, however, show a clear benefit from inserting the additional step. The mean total energy consumption decreases from $3.07$~kWh (baseline) to $2.57$~kWh (decorated), corresponding to a reduction of approximately $16\%$. Wasted energy during unoccupied periods is also reduced—from an average of $1.26$~kWh to $1.03$~kWh—representing an improvement of about $18\%$. These reductions confirm that the inserted step $N$ effectively suppresses unnecessary high-power operation when the room is empty. The cost of this improvement is minimal: the average occupied-period temperature error (ErrOcc) changes only marginally, from $6.73$ (baseline) to $6.79$ (decorated), indicating that temperature control during occupied periods is largely preserved.

The anomaly statistics further highlight the effect of DeMe. The baseline controller exhibits an average of $2.0$ environment anomalies per episode, with peaks of up to 8 anomalies (Episodes~10, 18, and~19). In contrast, the decorated controller reduces the mean EnvAnom to $0.65$, with a maximum of 3 anomalies in any episode. At the same time, step~$N$ actively intercepts anomalies, with N\_Anom ranging from 0 to 7 and an average of $2.25$ interventions per episode. Episodes~5, 11, 13, 18, and~19 show particularly frequent interventions, where the verification step repeatedly prevents high-power operation in unoccupied states.

\begin{figure*}[htbp]
    \centering
    \includegraphics[width=6.5in]{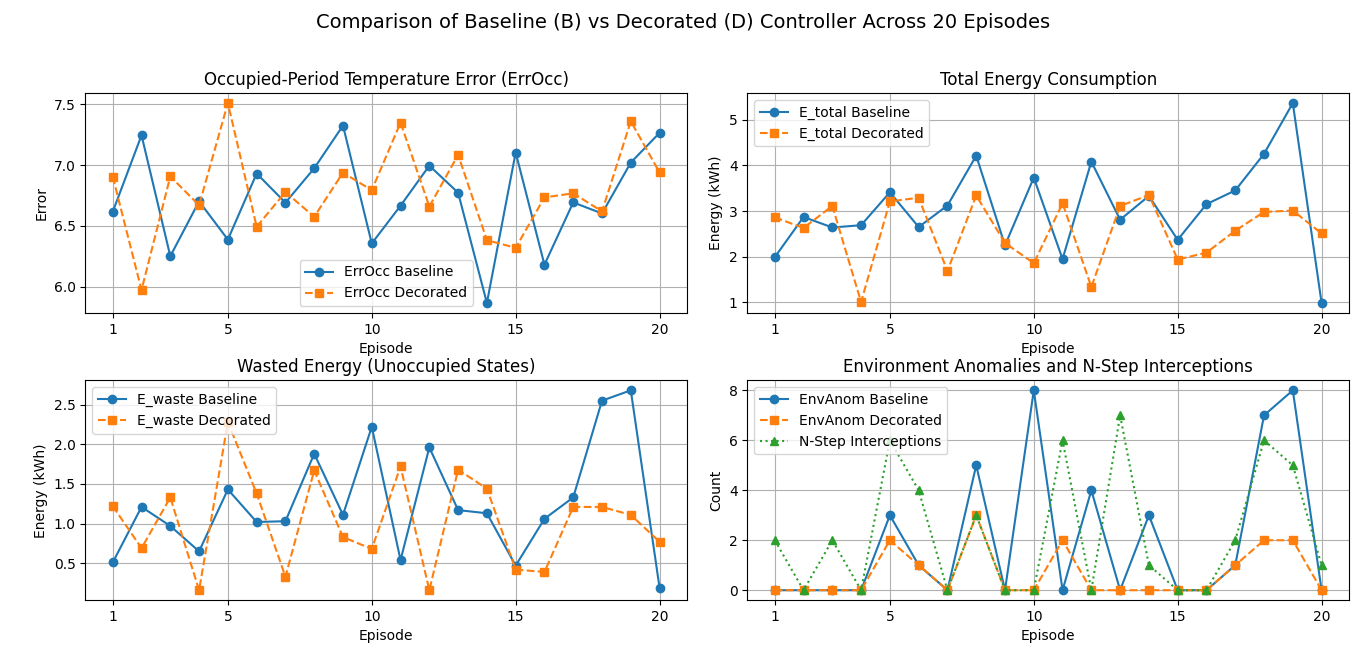}
    \caption{Per-episode comparison between the baseline controller (B) and the decorated controller with step~$N$ (D). The curves report occupied-period temperature error (ErrOcc), total energy consumption $E_{\text{total}}$, wasted energy $E_{\text{waste}}$, and anomaly counts (EnvAnom and N\_Anom) over 20 episodes.}
    \label{fig:dmom_stepN_compare}
\end{figure*}

Figure~\ref{fig:dmom_stepN_compare} visualizes ErrOcc, total energy, wasted energy, and anomaly counts over the 20 episodes. While the two controllers exhibit similar temperature-tracking behavior, the decorated controller consistently operates with lower total energy and reduced wasted energy, and it substantially suppresses anomalous ``no-person but high-power'' events.

Overall, these results highlight the core principle of DeMe:
\emph{by inserting an explicit verification step into the reasoning path, the controller adopts a more cautious and interpretable decision-making pattern, improving constraint awareness and reducing energy waste, while largely preserving tracking performance during occupied periods.}
Unlike post-processing or simple rule checks, DeMe modifies the generation process of the method itself, resulting in predictable and verifiable behavioral adjustments in dynamic IoT control scenarios.

% ===================== TABLE ==========================
\begin{table*}[htbp]
  \centering
  \caption{Per-episode comparison between baseline controller (B) and decorated controller with step~$N$ (D). ErrOcc: occupied-period temperature error; $E_{\text{total}}$: total energy; $E_{\text{waste}}$: wasted energy in unoccupied states; EnvAnom: environment anomalies; N\_Anom: anomalies intercepted by $N$.}
  \label{tab:dmom_stepN_episode_compare}
  \begin{tabular}{cccccccccc}
    \toprule
    Ep &
    ErrOcc$_B$ & ErrOcc$_D$ &
    $E_B$ & $E_D$ &
    $E_{\text{waste},B}$ & $E_{\text{waste},D}$ &
    EnvAnom$_B$ & EnvAnom$_D$ & N\_Anom$_D$ \\
    \midrule
    1  & 6.616 & 6.905 & 1.99 & 2.86 & 0.51 & 1.22 & 0 & 0 & 2 \\
    2  & 7.246 & 5.972 & 2.87 & 2.63 & 1.21 & 0.69 & 0 & 0 & 0 \\
    3  & 6.250 & 6.907 & 2.64 & 3.11 & 0.97 & 1.33 & 0 & 0 & 2 \\
    4  & 6.702 & 6.670 & 2.69 & 1.01 & 0.65 & 0.16 & 0 & 0 & 0 \\
    5  & 6.387 & 7.511 & 3.41 & 3.21 & 1.43 & 2.28 & 3 & 2 & 6 \\
    6  & 6.931 & 6.493 & 2.65 & 3.29 & 1.02 & 1.38 & 1 & 1 & 4 \\
    7  & 6.691 & 6.778 & 3.12 & 1.69 & 1.03 & 0.33 & 0 & 0 & 0 \\
    8  & 6.974 & 6.575 & 4.21 & 3.34 & 1.88 & 1.67 & 5 & 3 & 3 \\
    9  & 7.326 & 6.938 & 2.25 & 2.30 & 1.11 & 0.83 & 0 & 0 & 0 \\
    10 & 6.355 & 6.794 & 3.73 & 1.86 & 2.22 & 0.68 & 8 & 0 & 0 \\
    11 & 6.666 & 7.350 & 1.95 & 3.17 & 0.54 & 1.72 & 0 & 2 & 6 \\
    12 & 6.996 & 6.659 & 4.08 & 1.34 & 1.96 & 0.16 & 4 & 0 & 0 \\
    13 & 6.773 & 7.085 & 2.81 & 3.11 & 1.17 & 1.68 & 0 & 0 & 7 \\
    14 & 5.865 & 6.382 & 3.33 & 3.35 & 1.13 & 1.44 & 3 & 0 & 1 \\
    15 & 7.104 & 6.319 & 2.37 & 1.94 & 0.47 & 0.42 & 0 & 0 & 0 \\
    16 & 6.180 & 6.734 & 3.15 & 2.09 & 1.06 & 0.39 & 0 & 0 & 0 \\
    17 & 6.693 & 6.768 & 3.45 & 2.57 & 1.33 & 1.21 & 1 & 1 & 2 \\
    18 & 6.602 & 6.619 & 4.25 & 2.98 & 2.55 & 1.21 & 7 & 2 & 6 \\
    19 & 7.021 & 7.361 & 5.36 & 3.01 & 2.68 & 1.11 & 8 & 2 & 5 \\
    20 & 7.266 & 6.940 & 0.99 & 2.51 & 0.19 & 0.77 & 0 & 0 & 1 \\
    \bottomrule
  \end{tabular}
\end{table*}

\section{Conclusion} \label{sec:conclusion}

This paper introduced \emph{Method Decoration} (DeMe), a general framework that enhances the adaptability of LLM-driven IoT systems by explicitly modifying the method-generation path rather than altering the LLM itself. By incorporating decorations derived from hidden goals, accumulated learned methods, and environmental feedback, DeMe enables IoT agents to restructure their task-execution steps through pre-decoration, post-decoration, intermediate-step modification, and step insertion. Verification across two IoT-oriented scenarios—whole-process decoration for safety-aware reasoning and step-insertion decoration for HVAC control—demonstrated that decorations consistently improve method generation in dynamic IoT environments.

Future work will explore how DeMe can be scaled to more complex, multi-agent IoT systems and heterogeneous device networks. An important direction is developing automated mechanisms for selecting, weighting, and composing decorations when multiple forms of feedback or learned knowledge are available. We also plan to investigate how DeMe interacts with online learning, reinforcement learning, and continual learning frameworks to support long-term model evolution. Additionally, incorporating uncertainty quantification and trust-aware reasoning into the decoration process may further enhance safety in high-risk IoT settings. These extensions will help advance DeMe toward a fully autonomous, self-improving foundation for future intelligent IoT infrastructures.

% \section*{Acknowledgment}
% The authors thanks th.

% Can use something like this to put references on a page
% by themselves when using endfloat and the captionsoff option.
\ifCLASSOPTIONcaptionsoff
  \newpage
\fi

\bibliographystyle{IEEEtran}
\bibliography{ref}

% Generated by IEEEtran.bst, version: 1.14 (2015/08/26)
\begin{thebibliography}{10}
\providecommand{\url}[1]{#1}
\csname url@samestyle\endcsname
\providecommand{\newblock}{\relax}
\providecommand{\bibinfo}[2]{#2}
\providecommand{\BIBentrySTDinterwordspacing}{\spaceskip=0pt\relax}
\providecommand{\BIBentryALTinterwordstretchfactor}{4}
\providecommand{\BIBentryALTinterwordspacing}{\spaceskip=\fontdimen2\font plus
\BIBentryALTinterwordstretchfactor\fontdimen3\font minus
  \fontdimen4\font\relax}
\providecommand{\BIBforeignlanguage}[2]{{%
\expandafter\ifx\csname l@#1\endcsname\relax
\typeout{** WARNING: IEEEtran.bst: No hyphenation pattern has been}%
\typeout{** loaded for the language `#1'. Using the pattern for}%
\typeout{** the default language instead.}%
\else
\language=\csname l@#1\endcsname
\fi
#2}}
\providecommand{\BIBdecl}{\relax}
\BIBdecl

\bibitem{huang2025foundation}
J.~Huang, Y.~Xu, Q.~Wang, Q.~C. Wang, X.~Liang, F.~Wang, Z.~Zhang, W.~Wei,
  B.~Zhang, L.~Huang \emph{et~al.}, ``Foundation models and intelligent
  decision-making: Progress, challenges, and perspectives,'' \emph{The
  Innovation}, 2025.

\bibitem{gholami2024artificial}
H.~Gholami, ``Artificial intelligence techniques for sustainable reconfigurable
  manufacturing systems: An ai-powered decision-making application using large
  language models,'' \emph{Big Data and Cognitive Computing}, vol.~8, no.~11,
  p. 152, 2024.

\bibitem{su2025method}
H.~Su, ``Method-based reasoning for large language models: Extraction, reuse,
  and continuous improvement,'' \emph{arXiv preprint arXiv:2508.04289}, 2025.

\bibitem{qu2025mobile}
G.~Qu, Q.~Chen, W.~Wei, Z.~Lin, X.~Chen, and K.~Huang, ``Mobile edge
  intelligence for large language models: A contemporary survey,'' \emph{IEEE
  Communications Surveys \& Tutorials}, 2025.

\bibitem{yao2025efficient}
Z.~Yao, Y.~Xu, H.~Xu, Y.~Liao, and Z.~Xie, ``Efficient deployment of large
  language models on resource-constrained devices,'' \emph{arXiv preprint
  arXiv:2501.02438}, 2025.

\bibitem{zong2025integrating}
M.~Zong, A.~Hekmati, M.~Guastalla, Y.~Li, and B.~Krishnamachari, ``Integrating
  large language models with internet of things: applications,'' \emph{Discover
  Internet of Things}, vol.~5, no.~1, p.~2, 2025.

\bibitem{sarhaddi2025llms}
F.~Sarhaddi, N.~T. Nguyen, A.~Zuniga, P.~Hui, S.~Tarkoma, H.~Flores, and
  P.~Nurmi, ``Llms and iot: A comprehensive survey on large language models and
  the internet of things,'' \emph{Authorea Preprints}, 2025.

\bibitem{wei2022chain}
J.~Wei, X.~Wang, D.~Schuurmans, M.~Bosma, F.~Xia, E.~Chi, Q.~V. Le, D.~Zhou
  \emph{et~al.}, ``Chain-of-thought prompting elicits reasoning in large
  language models,'' \emph{Advances in neural information processing systems},
  vol.~35, pp. 24\,824--24\,837, 2022.

\bibitem{wang2022self}
X.~Wang, J.~Wei, D.~Schuurmans, Q.~Le, E.~Chi, S.~Narang, A.~Chowdhery, and
  D.~Zhou, ``Self-consistency improves chain of thought reasoning in language
  models,'' \emph{arXiv preprint arXiv:2203.11171}, 2022.

\bibitem{zhou2022least}
D.~Zhou, N.~Sch{\"a}rli, L.~Hou, J.~Wei, N.~Scales, X.~Wang, D.~Schuurmans,
  C.~Cui, O.~Bousquet, Q.~Le \emph{et~al.}, ``Least-to-most prompting enables
  complex reasoning in large language models,'' \emph{arXiv preprint
  arXiv:2205.10625}, 2022.

\bibitem{madaan2023self}
A.~Madaan, N.~Tandon, P.~Gupta, S.~Hallinan, L.~Gao, S.~Wiegreffe, U.~Alon,
  N.~Dziri, S.~Prabhumoye, Y.~Yang \emph{et~al.}, ``Self-refine: Iterative
  refinement with self-feedback,'' \emph{Advances in Neural Information
  Processing Systems}, vol.~36, pp. 46\,534--46\,594, 2023.

\bibitem{yao2022react}
S.~Yao, J.~Zhao, D.~Yu, N.~Du, I.~Shafran, K.~R. Narasimhan, and Y.~Cao,
  ``React: Synergizing reasoning and acting in language models,'' in \emph{The
  eleventh international conference on learning representations}, 2022.

\bibitem{schick2023toolformer}
T.~Schick, J.~Dwivedi-Yu, R.~Dess{\`\i}, R.~Raileanu, M.~Lomeli, E.~Hambro,
  L.~Zettlemoyer, N.~Cancedda, and T.~Scialom, ``Toolformer: Language models
  can teach themselves to use tools,'' \emph{Advances in Neural Information
  Processing Systems}, vol.~36, pp. 68\,539--68\,551, 2023.

\bibitem{dong2024self}
Y.~Dong, X.~Jiang, Z.~Jin, and G.~Li, ``Self-collaboration code generation via
  chatgpt,'' \emph{ACM Transactions on Software Engineering and Methodology},
  vol.~33, no.~7, pp. 1--38, 2024.

\bibitem{kamoi2024can}
R.~Kamoi, Y.~Zhang, N.~Zhang, J.~Han, and R.~Zhang, ``When can llms actually
  correct their own mistakes? a critical survey of self-correction of llms,''
  \emph{Transactions of the Association for Computational Linguistics},
  vol.~12, pp. 1417--1440, 2024.

\bibitem{mnih2015human}
V.~Mnih, K.~Kavukcuoglu, D.~Silver, A.~A. Rusu, J.~Veness, M.~G. Bellemare,
  A.~Graves, M.~Riedmiller, A.~K. Fidjeland, G.~Ostrovski \emph{et~al.},
  ``Human-level control through deep reinforcement learning,'' \emph{nature},
  vol. 518, no. 7540, pp. 529--533, 2015.

\bibitem{lillicrap2015continuous}
T.~P. Lillicrap, J.~J. Hunt, A.~Pritzel, N.~Heess, T.~Erez, Y.~Tassa,
  D.~Silver, and D.~Wierstra, ``Continuous control with deep reinforcement
  learning,'' \emph{arXiv preprint arXiv:1509.02971}, 2015.

\bibitem{chen2021iotrl}
A.~Uprety and D.~B. Rawat, ``Reinforcement learning for iot security: A
  comprehensive survey,'' \emph{IEEE Internet of Things Journal}, vol.~8,
  no.~11, pp. 8693--8706, 2020.

\bibitem{li2020iotadapt}
A.~S. Rajawat, S.~Goyal, C.~Chauhan, P.~Bedi, M.~Prasad, and T.~Jan,
  ``Cognitive adaptive systems for industrial internet of things using
  reinforcement algorithm,'' \emph{Electronics}, vol.~12, no.~1, p. 217, 2023.

\bibitem{sun2019edgelearning}
Q.~Liu, L.~Cheng, T.~Ozcelebi, J.~Murphy, and J.~Lukkien, ``Deep reinforcement
  learning for iot network dynamic clustering in edge computing,'' in
  \emph{2019 19th IEEE/ACM international symposium on cluster, Cloud and Grid
  Computing (CCGRID)}.\hskip 1em plus 0.5em minus 0.4em\relax IEEE, 2019, pp.
  600--603.

\bibitem{parisi2019continual}
G.~I. Parisi, R.~Kemker, J.~L. Part, C.~Kanan, and S.~Wermter, ``Continual
  lifelong learning with neural networks: A review,'' \emph{Neural networks},
  vol. 113, pp. 54--71, 2019.

\bibitem{ouyang2022instructgpt}
L.~Ouyang, J.~Wu, X.~Jiang, D.~Almeida, C.~Wainwright, P.~Mishkin, C.~Zhang,
  S.~Agarwal, K.~Slama, A.~Ray \emph{et~al.}, ``Training language models to
  follow instructions with human feedback,'' \emph{Advances in neural
  information processing systems}, vol.~35, pp. 27\,730--27\,744, 2022.

\bibitem{hu2022lora}
E.~J. Hu, Y.~Shen, P.~Wallis, Z.~Allen-Zhu, Y.~Li, S.~Wang, L.~Wang, W.~Chen
  \emph{et~al.}, ``Lora: Low-rank adaptation of large language models.''
  \emph{ICLR}, vol.~1, no.~2, p.~3, 2022.

\bibitem{lester2021prompttuning}
A.~Sabbatella, A.~Ponti, I.~Giordani, A.~Candelieri, and F.~Archetti, ``Prompt
  optimization in large language models,'' \emph{Mathematics}, vol.~12, no.~6,
  p. 929, 2024.

\bibitem{zaken2022bitfit}
E.~B. Zaken, Y.~Goldberg, and S.~Ravfogel, ``Bitfit: Simple parameter-efficient
  fine-tuning for transformer-based masked language-models,'' in
  \emph{Proceedings of the 60th Annual Meeting of the Association for
  Computational Linguistics (Volume 2: Short Papers)}, 2022, pp. 1--9.

\bibitem{lewis2020rag}
P.~Lewis, E.~Perez, A.~Piktus, F.~Petroni, V.~Karpukhin, N.~Goyal,
  H.~K{\"u}ttler, M.~Lewis, W.-t. Yih, T.~Rockt{\"a}schel \emph{et~al.},
  ``Retrieval-augmented generation for knowledge-intensive nlp tasks,''
  \emph{Advances in neural information processing systems}, vol.~33, pp.
  9459--9474, 2020.

\bibitem{borgeaud2022retro}
J.~Lan, J.~Chen, Z.~Liu, C.~Li, S.~Bao, and D.~Lian, ``Retro*: Optimizing llms
  for reasoning-intensive document retrieval,'' \emph{arXiv preprint
  arXiv:2509.24869}, 2025.

\bibitem{wei2021finetuned}
J.~Wei, M.~Bosma, V.~Y. Zhao, K.~Guu, A.~W. Yu, B.~Lester, N.~Du, A.~M. Dai,
  and Q.~V. Le, ``Finetuned language models are zero-shot learners,''
  \emph{arXiv preprint arXiv:2109.01652}, 2021.

\bibitem{kojima2022zero_shot}
T.~Kojima, S.~S. Gu, M.~Reid, Y.~Matsuo, and Y.~Iwasawa, ``Large language
  models are zero-shot reasoners,'' \emph{Advances in neural information
  processing systems}, vol.~35, pp. 22\,199--22\,213, 2022.

\bibitem{brown2020gpt3}
B.~Mann, N.~Ryder, M.~Subbiah, J.~Kaplan, P.~Dhariwal, A.~Neelakantan,
  P.~Shyam, G.~Sastry, A.~Askell, S.~Agarwal \emph{et~al.}, ``Language models
  are few-shot learners,'' \emph{arXiv preprint arXiv:2005.14165}, vol.~1,
  no.~3, p.~3, 2020.

\bibitem{su2025difference}
H.~Su, ``Difference-guided reasoning: A temporal-spatial framework for large
  language models,'' \emph{arXiv preprint arXiv:2509.20713}, 2025.

\end{thebibliography}

% biography section
%
% If you have an EPS/PDF photo (graphicx package needed) extra braces are
% needed around the contents of the optional argument to biography to prevent
% the LaTeX parser from getting confused when it sees the complicated
% \includegraphics command within an optional argument. (You could create
% your own custom macro containing the \includegraphics command to make things
% simpler here.)
%\begin{IEEEbiography}[{\includegraphics[width=1in,height=1.25in,clip,keepaspectratio]{mshell}}]{Michael Shell}
% or if you just want to reserve a space for a photo:

\begin{IEEEbiography}{Hong Su}
  received the MS and PhD degrees, in 2006 and 2022, respectively, from Sichuan University, Chengdu, China. He is currently a researcher of Chengdu University of Information Technology Chengdu, China. His research interests include blockchain, cross-chain and smart contract.
\end{IEEEbiography}

% You can push biographies down or up by placing
% a \vfill before or after them. The appropriate
% use of \vfill depends on what kind of text is
% on the last page and whether or not the columns
% are being equalized.

%\vfill

% Can be used to pull up biographies so that the bottom of the last one
% is flush with the other column.
%\enlargethispage{-5in}

% that's all folks
\end{document}